\documentclass{article}

\PassOptionsToPackage{numbers, compress}{natbib}

\usepackage[preprint]{neurips_2024}

\usepackage[preprint]{neurips_2024}

\usepackage{graphicx} 
\usepackage[utf8]{inputenc} 
\usepackage[T1]{fontenc}    
\usepackage{hyperref}       
\usepackage{url}            
\usepackage{booktabs}       
\usepackage{amsfonts}       
\usepackage{nicefrac}       
\usepackage{microtype}      
\usepackage{xcolor}         
\usepackage{float}
\usepackage{tcolorbox} 
\usepackage{comment}
\usepackage{subcaption}
\usepackage{enumitem}
\usepackage{tocloft}   
\title{LLM Embeddings for Deep Learning on Tabular Data}

\author{%
  Boshko Koloski\textsuperscript{1}\thanks{Equal contribution.}\space Andrei Margeloiu$^*$\textsuperscript{2} \space Xiangjian Jiang\textsuperscript{2}$^*$ \\ \textbf{Bla\v{z} \v{S}krlj\textsuperscript{1} \space Nikola Simidjievski\textsuperscript{2,3} \space Mateja Jamnik\textsuperscript{2}} \\
  \textsuperscript{1} Jo\v{z}ef Stefan Institute and Postgraduate School, SI \\
  \textsuperscript{2}Department of Computer Science and Technology, University of Cambridge, UK \\
  \textsuperscript{3}PBCI, Department of Oncology, University of Cambridge, UK \\
  \texttt{\{boshko.koloski, blaz.skrlj@ijs.si\}} \\ 
  \texttt{\{am2770, xj265, ns779, mj201\}@cam.ac.uk} \\
}

\begin{document}

\maketitle
    
\begin{abstract}    

    Tabular deep-learning methods require embedding numerical and categorical input features into high-dimensional spaces before processing them. Existing methods deal with this heterogeneous nature of tabular data by employing separate type-specific encoding approaches. This limits the cross-table transfer potential and the exploitation of pre-trained knowledge.
We propose a novel approach that first transforms tabular data into text, and then leverages pre-trained representations from LLMs to encode this data, resulting in a plug-and-play solution to improving deep-learning tabular methods. We demonstrate that our approach improves accuracy over competitive models, such as MLP, ResNet and FT-Transformer, by  validating on seven classification datasets.
\end{abstract}

\addtocontents{toc}{\protect\setcounter{tocdepth}{0}}

\section{Introduction and Background}

Tabular data is common in domains such as education \cite{higher_education_students_performance_evaluation_856}, banking \cite{bank_marketing_222,statlog_(german_credit_data)_144} and medicine \cite{heart_disease_45,hepatitis_46,blood_transfusion_service_center_176}.
Thus, in recent years, tabular deep learning for predictive tasks has gained traction, with many architectures achieving state-of-the-art results \cite{ye2024closer}. A key challenge in applying neural networks to tabular data is determining \textit{how to effectively represent the features}. Tabular data is typically represented with a mix of numerical and categorical features, living in high dimensions~\cite{koloski2023latent,margeloiu2024gcondnet}. 
A common approach to effectively representing features involves
encoding numerical features differently from categorical ones~\cite{fttransformer,arik2021tabnet,tpberta,thielmann2024mambular}. Once separated, various representation learning techniques are employed \cite{bengio2013representation}. For example, numerical features are often transformed and then encoded using linear transformations, while categorical features are typically represented with lookup embeddings \cite{borisov2110deep,gorishniy2022embeddings,wu2024deep}. These embeddings provide a foundation for architectures that ensure permutation invariance at the feature level, such as transformers \cite{vaswani}. However, this approach is limiting because it separates features and trains individual embeddings from scratch. This is particularly restrictive for domains where data is scarce and usually  consists of mixed numerical and categorical features. 


TabPFN~\cite{hollmann2022tabpfn} is a pre-trained tabular transformer trained on synthetic classification datasets, achieving competitive results and learning useful representations for various downstream predictors~\citep{margeloiu2024tabmda}. Despite its success, TabPFN has limited adaptability to real-world datasets, it only supports numerical data, and is constrained by the number of features and samples it can handle. Recently, TPBERTa\cite{tpberta}  was proposed as an approach where features are first encoded separately based on their type, and then the RoBERTa \cite{liu2019roberta} language model is fine-tuned across datasets. This method shows promising potential of utilising underlying LLM as a backbone, however it requires prior separate processing of numerical and categorical features, as well as tuning a relatively big model, which is limited by the number of features it can handle. Zhang et al.~\cite{zhang2023generativetablepretrainingempowers} showed the benefit of transforming tabular data to text, and learning representations by auto-regressively pre-training an LLM for downstream tasks.
TabLLM~\cite{hegselmann2023tabllm} demonstrates that LLMs can be fine-tuned for downstream tabular tasks, achieving competitive results in few-shot settings. However, this approach is computationally expensive for larger datasets and limited by the number of features it can handle, as the context window of LLMs is constrained. 
The majority of proposed deep-learning tabular approaches disentangle the feature names from their values when encoding, thus restricting the model's ability to learn meaningful interactions between them. On the other hand, representations acquired from LLMs, have shown great potential for retrieval and classification tasks~\cite{muennighoff-etal-2023-mteb} even for tabular data~\cite{luo2024bgelandmarkembeddingchunkingfree}. This opens the question: \emph{Can LLM representations be effectively utilised as feature encoders within current high-performant tabular deep-learning frameworks?}

\begin{figure}[t]
    \centering
    \includegraphics[trim={20pt 50pt 20pt 10pt},clip, width=\textwidth]{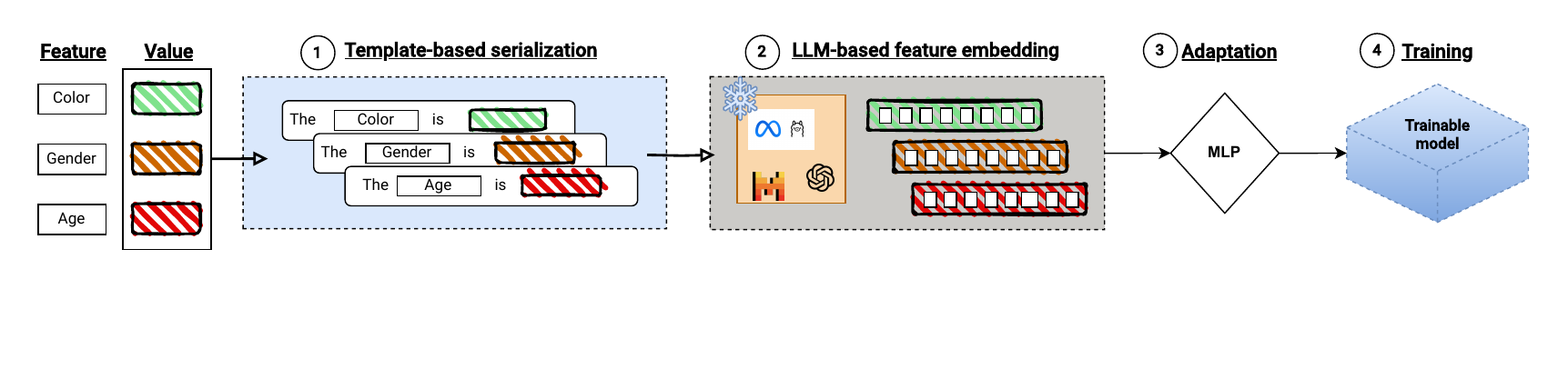}
    \caption{Schema of our proposed methodology. (1)~The input is first serialised, feature by feature, into sentences. (2)~Large Language Models (LLMs) are used to extract embeddings of the inputs. (3)~We project and adapt the embeddings with an MLP. (4)~We apply trainable models that utilise the LLM embeddings for feature encoding.}
    \label{fig:modus_operandi}
\end{figure}

\looseness-1
To answer this question, in this paper we investigate the capabilities of recent LLM models as a new, training-free representation-learning method for tabular data. Our approach (see Figure \ref{fig:modus_operandi}) represents each feature and its value interactions as a sentence, which is subsequently encoded by an LLM \cite{behnamghader2024llm2vec}, enabling feature-to-value interaction, regardless of the heterogeneity of features. We demonstrate that these embeddings, without additional fine-tuning of the LLM, can enhance the performance of state-of-the-art models like FT-Transformer, and enable the transfer of previously acquired LLM knowledge.

The main contribution of this work is a new LLM-based encoding method of tabular data, that operates effectively regardless of the feature type or the domain from which the data originates. This method can be easily integrated with any model that uses input-feature embeddings, enhancing downstream results. We demonstrate that the method performs well regardless of the chosen LLM and can be applied to diverse classification datasets and different learners.

\section{Methodology}

\textbf{Formal definition.} Let $D$ be a tabular dataset consisting of $N$ samples and $M$ features describing the input matrix $X$, used to predict the target vector $y$, where $(X_{i}, y_{i})$ denotes the input features and the label of the $i$-th row. Let $g$ be an LLM-embedder that maps each feature $m \in M$ to a $d$-dimensional space $\mathbb{R}^{M \times d}$. By mapping the whole input $X$, we obtain the embedded matrix $E(X) \in \mathbb{R}^{N \times M \times d}$. Finally, a trainable model $f$ (e.g., FT-Transformer) is trained on the embedded data to learn the mapping from the embedded input to the output, $f(\theta; E) \mapsto \hat{y}$.


\textbf{Feature Encoding with LLMs.} We use LLMs as the encoding function $g$, which maps the input feature values to the LLM's embedding space. In particular, we first serialise the features into text and then embed the serialised text through the LLM to generate LLM-based embeddings. Previous work has shown that different prompting strategies have minimal impact~\cite{hegselmann2023tabllm}. Following this, we serialise the features using the template: \colorbox{lightgray}{\texttt{This \{col\} is \{value\}.}}\footnote{Examples of serialisation are shown in Appendix~\ref{sec:examples}.} 

\textbf{Embedding adaptation.} Next, we apply a shallow, one-layer neural network on top of the obtained LLM representations to project the LLM embeddings as inputs for the trainable model. The motivation behind this is twofold: (a)~To ensure a fair comparison between models by projecting all embeddings to the same dimension, thus forcing the models to operate within the same dimensionality. (b)~While the base variant models update their randomly initialised layers by fitting to the data, this approach introduces a similar adaptability mechanism to the frozen LLM embeddings, ensuring that they can be aligned with the downstream task.

\textbf{Trainable Models.} Recent work~\cite{grinsztajn2022tree} indicates that approaches like ResNet and FT-Transformer generally provide stable and strong performance on various tabular datasets. Therefore, in this paper, we evaluate these methods alongside a baseline MLP, using the same model architectures across all experiments while focusing solely on input embeddings:
\begin{itemize}[leftmargin=15pt, topsep=0pt]
    \item ResNet: Three layered variant with layers 256, 128 and 32 neurons; with Skip connections and SELU activation function, similar as in \cite{thielmann2024mambular}; batch norm is applied. 
    \item Multi-layer Perceptron (MLP): Similarly, 3 layered MLP with 256, 128 and 32 neurons. 
    \item FT-Transformer \cite{fttransformer}: We employ the FT-transformer variant with 4 layers and 8 heads, aggregating on the `CLS' token. We exclude other multi-head attention and transformer variants, such as the TabTransformer model, as they become equivalent to the FT-Transformer when all features are encoded uniformly.    
\end{itemize}


\section{Evaluation} We assess the impact of LLM-based embeddings on the classification accuracy on downstream tasks. To this end, we select seven datasets from three different domains: general, banking, and medical. Further details about the datasets can be found in Appendix~\ref{sec:dataset_info}. To avoid contamination, we select the BGE~\cite{luo2024bgelandmarkembeddingchunkingfree} embedding\footnote{\url{https://huggingface.co/BAAI/bge-base-en-v1.5}} as the LLM backbone representation, since it performs well on the MTEB benchmark~\cite{muennighoff-etal-2023-mteb}, and works well for tabular data retrieval \cite{khanna2024tabular}. BGE was trained on text-only input formats without using any training steps, where data was translated from other formats (e.g., tables) into text to enhance the language modelling process.

We address three \textbf{research questions}: \textbf{(a)}~Do LLM-based embeddings improve the performance of trainable predictors on downstream tasks? \textbf{(b)}~Are LLM embeddings effective only across specific domains, or do they perform well in general contexts? \textbf{(c)}~How does the choice of LLM affect the downstream performance?

\looseness-1
\par \textbf{Experimental Setup.} We evaluate the impact of training tabular neural networks with and without LLM-based embeddings. We fix the dimensionality of the embedding layer to 1024, independent of the model architecture, to ensure fairness. The weights of the LLM are frozen during model training. We perform 10 random train/test splits (with different random seeds), using a 70/30 stratified split of the data, and report the accuracy on the test set. We use 20\% of the training split for validation. Models are trained for up to 100 epochs, with early stopping set to a patience of 10 and a minimum improvement delta of~0.01 in terms of validation loss.

\section{Results and Discussion}
Table \ref{tab:tab1} presents the results from the comparison between models with LLM embeddings (denoted as ``with LLM'') and the base variants (``base''). We find that, on average, the LLM-based embeddings produce better results, outperforming the base trainable models.

\begin{table}[h]
\centering
\caption{Impact of the LLM embeddings, compared to the base versions of the DL models.}
\resizebox{0.9\textwidth}{!}{\begin{tabular}{l |cc| cc| cc}
\toprule
 & \multicolumn{2}{c|}{\textbf{ResNet}}  & \multicolumn{2}{c|}{\textbf{MLP}}  & \multicolumn{2}{c}{\textbf{FT-Transformer}}\\
Dataset & with LLM & base  & with LLM & base &  with LLM & base  \\
\midrule

blood transfusion & 74.36 ± 1.99 & \textbf{75.16 ± 1.84} & 75.42 ± 1.17 & \textbf{76.53 ± 1.51} & 75.20 ± 1.06 & \textbf{76.00 ± 0.01} \\
credit-g &  \textbf{74.00 ± 1.66} & 70.50 ± 0.81 & \textbf{74.20 ± 1.75} & 70.00 ± 0.00 &\textbf{ 73.63 ± 1.74} & 73.37 ± 1.23 \\
bank-marketing  &\textbf{ 90.08 ± 0.23 }& 89.72 ± 0.39 & \textbf{90.10 ± 0.16} & 88.42 ± 0.17 &\textbf{ 90.33 ± 0.21} & 89.14 ± 0.21 \\
diabetes & \textbf{72.33 ± 2.35} & 71.35 ± 2.95 & \textbf{73.00 ± 1.85} & 68.52 ± 3.02 & \textbf{71.51 ± 2.20} & 65.25 ± 2.93 \\
heart & 57.80 ± 2.21 & \textbf{59.34 ± 1.47}   & \textbf{58.35 ± 2.66} & 54.61 ± 1.47 &\textbf{ 59.89 ± 3.32 }& 57.80 ± 2.21 \\
hepatits  & \textbf{79.78 ± 4.04} & 76.59 ± 3.47 & \textbf{80.00 ± 3.64} & 78.51 ± 2.54 & \textbf{80.43 ± 2.98} & 78.72 ± 0.01   \\
student-performance  & \textbf{22.50 ± 4.47} & 21.36 ± 3.89 & \textbf{22.73 ± 4.14} & 20.91 ± 4.76 & \textbf{23.63 ± 6.70} & 20.91 ± 6.32   \\
\midrule
Average & \textbf{67.98} & 66.43 & \textbf{67.69} & 65.22 & \textbf{68.09} & 65.03 \\

\bottomrule
\end{tabular}}
\label{tab:tab1}

\end{table}

Model-wise, the largest gain is for the FT-Transformer (3.05\%), which is not surprising, since the transformer mechanism enables powerful interactions between the LLM embedded features to be learned. Dataset wise, we only notice decrease in performance for the `blood transfusion', where the base models outperforms the LLM variant for all three learners. We hypothesise that this is due to the specificity of the dataset, as all four features are numerical, and the domain is niche, not sufficiently covered by a general-purpose LLM such as BGE. Moreover, representing numerical data has been shown to be a bottleneck for general reasoning tasks~\cite{schwartz2024numerologicnumberencodingenhanced}, which could have impacted the results. 
On the other hand, for datasets where categorical features are widely present, such as `credit-g' and `student-performance', we observe consistent performance gains, suggesting that DL methods can significantly benefit from contextualising input by injecting feature-to-value interactions via LLMs. Additionally, we note that for datasets where feature names are descriptive and do not contain dashes, such as the medical `hepatitis' datasets, our approach can outperform state-of-the-art methods.

\looseness-1
Next, we investigate \textit{how the choice of an LLM impacts the downstream performance}. We select three datasets where the LLMs have varying impact compared to the base models: `heart', `hepatitis', and `student-performance'. Our goal is to determine if the performance improvement depends on the size and capabilities of the LLM. We investigate several variants, including a decoder-based LLaMa3 embeddings~\cite{dubey2024llama,behnamghader2024llm2vec} (with 4096 dimensions) and an encoder-based mini-LM \cite{minim-10957} (with 386 dimensions).

\begin{figure}[t]
    \centering
    \includegraphics[width=0.85\linewidth]{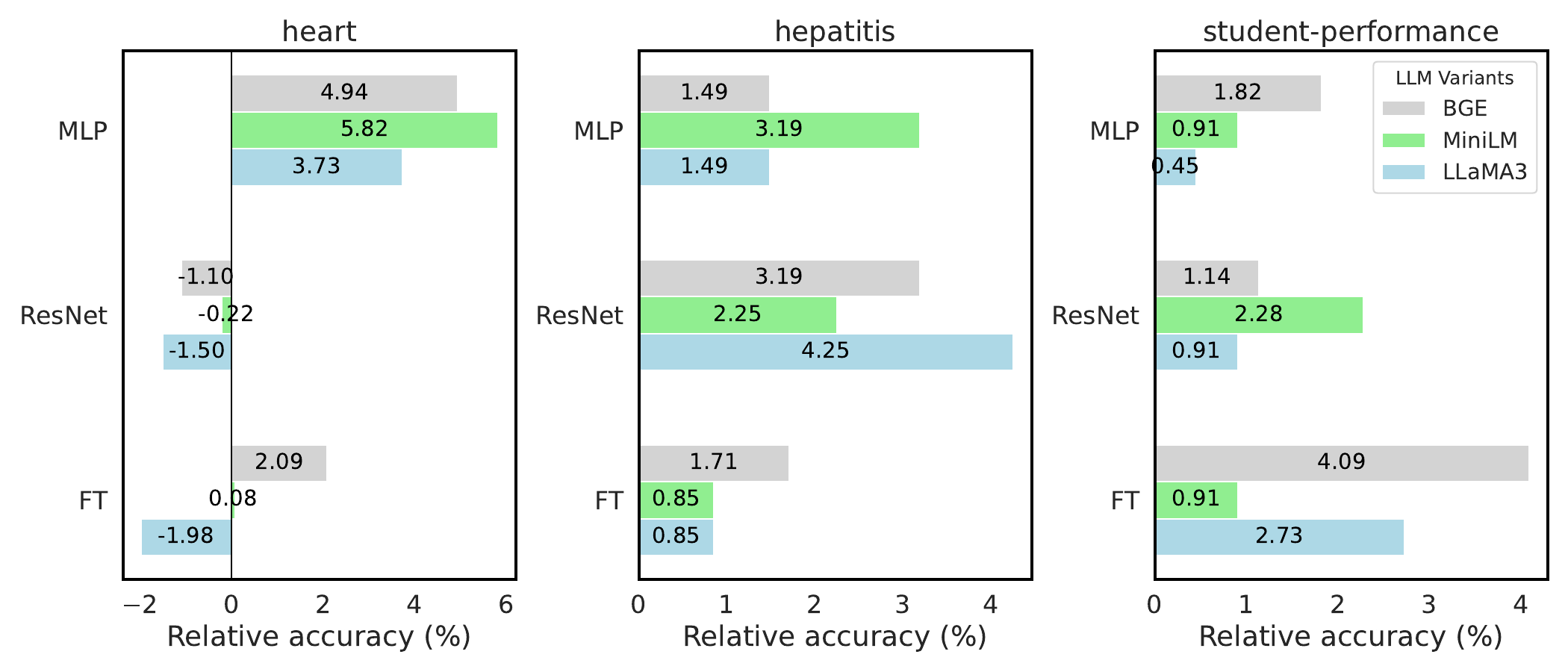}
    \caption{Comparing relative test performance of base models and their LLM-enhanced variants. Using LLMs generally improves performance, with BGE showing the most consistent improvements.}
    \label{fig:ablation1-llm-selection}
\end{figure}

Figure~\ref{fig:ablation1-llm-selection} suggests that selecting larger LLMs, such as LLama3, generally leads to improved performance. However, in some cases, such as the `heart' dataset, where feature names are highly specific, the model's performance can deteriorate. These findings imply that using bigger models, trained on more general sources, may yield better results for downstream tasks. However, we find that model selection tends to influence the magnitude of the performance change rather than the direction, with general trends prevailing across different models, on the tested datasets. We observe that the smaller BGE model represents a good choice overall for downstream tasks. As with other tasks~\cite{dong2024generalization}, one needs to be careful when selecting and benchmarking LLMs on tabular tasks, as they might be contaminated during pre-training~\cite{bordt2024elephants} by using publicly available data, such as some of the ones considered in this work (e.g., the `heart' dataset~\cite{heart_disease_45} from the UCI repository). \looseness=-1

Finally, we compare the LLM-enhanced models with competitive machine-learning baselines (see Table 2 in the Appendix~\ref{sec:hyperpar}). We find that the LLM-enhanced models (with BGE), while having an improved performance, struggle to outperform ensemble models such as Random Forests and XGBoost on these tasks. This is inline with similar findings reported in related work \cite{hegselmann2023tabllm,grinsztajn2022tree}, showing that strong tree-based ensemble learners can still outperform tabular network models. Nevertheless, we note that in all cases but one (`blood transfusion'), our proposed LLM embedding-based models manage to further narrow this gap and produce a comparable performance to tree-based learners.


\section{Conclusion and Further Work}
Our results demonstrate that LLM embeddings can: \textbf{(a)}~enhance the performance of highly effective tabular deep-learning methods (refer to Appendix~\ref{sec:stat} for statistical tests), \textbf{(b)}~be applied across a wide range of domains, and \textbf{(c)}~leverage various LLMs as embedders, with even smaller models like BGE showing strong performance. While the embeddings improve base model performance and narrow the gap, they generally still fall behind non-neural models (more discussion of the limitations is in Appedix~\ref{sec:limitation}). We further demonstrate in Appendix~\ref{sec:viz} that the proposed embeddings effectively capture meaningful properties by semantically grouping various feature-to-value interactions. In the future, we plan to extend our approach to cross-table training and explore how LLMs can better enhance feature-to-value interactions. \looseness=-1

\bibliography{bibliograph}
\bibliographystyle{plain}

\clearpage

\appendix
\addtocontents{toc}{\protect\setcounter{tocdepth}{2}}

\tableofcontents

\section{Limitations}
\label{sec:limitation}
A limitation of the proposed approach is its increased computational complexity. As each feature-value pair requires querying an additional LLM, complexity scales linearly with the number of features. While embeddings for finite-set features can be cached, continuous inputs require repeated querying, adding a persistent computational load. The results show that the choice of language model might influence downstream performance, contributing to the overall complexity. However, even small LLMs, tuned for embedding textual data, perform well, with BGE embeddings yielding strong results across tasks. Another limitation arises when feature names lack clear descriptions or are domain-specific and not well-covered by the LLM. To address this, integrating formal knowledge from knowledge bases, as suggested by \cite{ruiz2023enabling}, may provide a solution.

\section{Related work}

\textbf{Feature Embeddings} Embedding features is usually approached depending on the type of feature. Numerical features have traditionally been either transformed using a linear model \cite{hollmann2022tabpfn,fttransformer} or discretised by a variation of binning techniques \cite{tpberta,huang2020tabtransformer}. Binning as a technique for obtaining a pre-trained tabular deep learning model has shown promising results \cite{lee2023binning}. Categorical features are typically embedded via a lookup embedding layer \cite{fttransformer,huang2020tabtransformer}, with various techniques influencing their use \cite{borisov2110deep,gorishniy2022embeddings,wu2024deep}. Recently, tree-based feature embeddings have shown promising potential \cite{tpberta,li2023treeregularized}. Parallel to our work, \cite{tpberta} explored using word embeddings for embedding only feature names and categorical inputs, though they disentangled and embedded each word separately. To our knowledge, we are the first to explore LLM-based embeddings, both encoder- and decoder-based, for embedding of tabular data.

\textbf{LLMs and Tabular Data} Learning on serialised tabular data to text has been prominent in mining relational tabular data \cite{PEROVSEK20156442,lavravc2020propositionalization}. With the introduction of pre-trained models, a plethora of tabular models for table understanding and downstream prediction have been proposed, as shown in the recent survey \cite{fang2024large}. The majority of these applications focus on serialising the inputs and fine-tuning large language models for prediction \cite{hegselmann2023tabllm,slack2023tabletlearninginstructionstabular,zhang2023generativetablepretrainingempowers,dinh2022lift}. Another line of work focuses on using LLMs as classifiers, where inputs are tokenised and mapped to the LLM vocabulary \cite{tpberta}. The idea of leveraging the potential of LLMs for transfer-learning across tables and feature encoding was shown as useful in ~\cite{wang2022transtab}, however they propose different encoding for different input types adding a complexity by design. Recently, focus to incorporate LLM priors by prompting them to order the input features and this was exploited by traditional ML models \cite{zhu2023incorporatingllmpriorstabular} showing promising results. LLMs showed remarkable potential as feature engineers as well \cite{han2024large}. However, using LLMs in this manner is either computationally heavy e.g. fine-tuning or requires careful prompt creation which can be laborious. 

\textbf{LLMs and Text Embeddings} Semantically embedding texts is one of the main tasks of interest in NLP, resulting in a benchmarking effort called the Massive Text Embedding Benchmark (MTEB) \cite{muennighoff-etal-2023-mteb}. Traditionally, encoder-only model variants like sentence-based BERT \cite{reimers-2019-sentence-bert} were popular, with variants like BGE performing among the top performers \cite{luo2024bgelandmarkembeddingchunkingfree}. Recently, focus has shifted towards extracting embeddings from LLMs due to their remarkable capabilities \cite{kojima2022large}, where internal model embeddings, e.g., from the LLaMa3 model \cite{dubey2024llama}, are extracted with \cite{behnamghader2024llm2vec} or without tuning \cite{jiang2023scalingsentenceembeddingslarge} for document representation. Notably, the tokenisation process of LLMs, based on information-theoretic principles with byte-pair encoding techniques prevailing~\cite{bpe_suboptimal}, has been shown to negatively impact numerical reasoning tasks~\cite{schwartz2024numerologicnumberencodingenhanced}.

\section{Availability and Reproducibility} All experiments were conducted in a Singularity environment, and measurements were tracked using Weights and Biases (WandB) to ensure reproducibility of the results. 

\section{Datasets}
\label{sec:dataset_info}
We identified several different datasets based on the number of inputs and their relevance to the related work. We selected datasets from various domains, such as medical and financial, to assess how well the general-domain LLM embeddings can contribute to both general features like those in banking and more specific features such as those in medicine. Each dataset is split into a 70/30 stratified train-test split, and for early stopping, we use 20\% of the training data. We use the same splits across methods, to ensure for reproducibility and fairness of evaluation. The following datasets are used:

\begin{itemize}
    \item \textbf{bank-marketing}~\cite{bank_marketing_222}: A marketing dataset from a Portuguese bank, where the task is to determine if a customer subscribes to a deposit.
    \item \textbf{credit-g}~\cite{statlog_(german_credit_data)_144}: A banking dataset where the goal is to predict whether a given customer has good or bad credit risk for credit allowance.
    \item \textbf{heart} \cite{heart_disease_45}: A dataset where the goal is to predict the presence of heart disease and its severity, ranging from not present (0) to full presence (4), based on 14 clinical measurements.
    \item \textbf{diabetes} \cite{diabetes_34}: A dataset where the goal is to predict whether a patient has diabetes or not.
    \item \textbf{hepatitis}~\cite{hepatitis_46}: A dataset aimed at predicting whether a patient will live or die, given 19 features.
    \item \textbf{blood transfusion}\cite{blood_transfusion_service_center_176}: A dataset for churn prediction of Taiwanese blood donors, based on four numerical values.
    \item \textbf{student-performance}~\cite{higher_education_students_performance_evaluation_856}: A dataset for prediction of students performance ranked in 7 levels from failure (0) to excellent (7). 

\end{itemize}

\begin{table}[H]
\caption{Dataset Characteristics}

\label{tab:dataset_info}
\resizebox{\textwidth}{!}{\begin{tabular}{lr|rrrrr|rrr}
\toprule
Dataset & Domain & \#N & \#F & \#N/\#F & \#Cat. & \#Num. & Classes & Min Support & Max Support \\
\midrule
bank-marketing & banking & 45211 & 16 & 2825.69 & 9 & 7 & 2 & 11.70\% & 88.30\% \\
credit-g & banking & 1000 & 20 & 50.00 & 13 & 7 & 2 & 30.00\% & 70.00\% \\ \midrule
heart & medicine & 303 & 13 & 23.31 & 0 & 13 & 5 & 4.29\% & 54.13\% \\
hepatitis & medicine & 155 & 19 & 8.16 & 0 & 19 & 2 & 20.65\% & 79.35\% \\
diabetes & medicine & 768 & 8 & 96.00 & 0 & 8 & 2 & 34.90\% & 65.10\% \\
blood-transfusion & medicine & 748 & 4 & 187.00 & 0 & 4 & 2 & 23.80\% & 76.20\% \\ \midrule
student-performance & academics & 145 & 31 & 4.68 & 29 & 2 & 8 & 5.52\% & 24.14\% \\
\bottomrule
\end{tabular}}
\end{table}

\section{Comparison of the LLM-enhanced models to Strong Baselines}
\label{sec:hyperpar}

To further assess the performance of the LLM-enhanced models, with respect to standard approaches for modelling tabular data, we select three strong tabular data learners as per \cite{grinsztajn2022tree}: Stochastic-Gradient Descent (SGD), Random-Forest (RF) and eXtreme Gradient Boosting (XGB). We fine-tune each with a range of hyperparameters tuned through grid search on the 20\% of the training data, for each repetition.

\begin{itemize}
    \item \textbf{Random Forest (RF)}: \texttt{RandomForestClassifier} with hyperparameters:
    \begin{itemize}
        \item \texttt{n\_estimators}: [50, 100, 200]
        \item \texttt{max\_depth}: [None, 10, 20]
        \item \texttt{min\_samples\_split}: [2, 5, 10]
    \end{itemize}
    
    \item \textbf{XGBoost (XGB)}: \texttt{XGBClassifier} with hyperparameters:
    \begin{itemize}
        \item \texttt{n\_estimators}: [50, 100, 200]
        \item \texttt{learning\_rate}: [0.01, 0.1, 0.2]
        \item \texttt{max\_depth}: [3, 5, 7]
    \end{itemize}
    
    \item \textbf{SGD}: \texttt{SGDClassifier} with hyperparameters:
    \begin{itemize}
        \item \texttt{loss}: ['hinge', 'log\_loss']
        \item \texttt{alpha}: [0.0001, 0.001, 0.01]
        \item \texttt{max\_iter}: [1000, 2000]
        \item \texttt{tol}: [1e-3, 1e-4]
    \end{itemize}
\end{itemize}

The results in Table \ref{tab:res2} show that non-neural gradient-boosting trees, still outperform neural models, as found in \cite{grinsztajn2022tree,hegselmann2023tabllm}.
\begin{table}[H]
\caption{Comparison of the ResNet, MLP and FT models powered by the LLM embeddings and three strong baselines. Results in bold indicate the best-performance per dataset.}
\resizebox{\textwidth}{!}{\begin{tabular}{llll|lll}
\toprule
dataset & SGD & XGB & RF & ResNet & MLP & FT \\
\midrule
blood transfusion & 58.44 ± 18.46 & \textbf{76.89 ± 2.55} & 76.84 ± 2.37 & 74.36 ± 1.99 & 75.42 ± 1.17 & 75.20 ± 1.06 \\
credit-g & 58.87 ± 18.05 & 74.93 ± 1.98 & \textbf{75.27 ± 1.38} & 74.00 ± 1.66 & 74.20 ± 1.75 & 73.63 ± 1.74 \\
bank-marketing  & 75.31 ± 16.93 & \textbf{90.77 ± 0.24} & 90.52 ± 0.13 & 90.08 ± 0.23 & 90.10 ± 0.16 & 90.33 ± 0.21 \\
heart  & 47.03 ± 5.62 & 55.60 ± 2.89 & 57.25 ± 2.61 & 57.80 ± 2.50 & 58.35 ± 2.66 & \textbf{59.89 ± 3.33} \\
hepatitis & 77.02 ± 3.30 & 81.28 ± 3.86 & \textbf{83.19 ± 3.24} & 79.79 ± 4.04 & 80.00 ± 3.64 & 80.43 ± 2.98 \\
diabetes  & 58.96 ± 11.38 & 73.64 ± 1.91 & \textbf{75.11 ± 2.48} & 72.33 ± 2.35 & 73.00 ± 1.85 & 71.51 ± 2.20 \\
student-performance  & 20.91 ± 5.34 & \textbf{28.41 ± 4.04} & \textbf{28.41 ± 4.04} & 22.50 ± 4.48 & 22.73 ± 4.15 & 23.64 ± 6.71 \\
\bottomrule
\end{tabular}}
\label{tab:res2}
\end{table}

\section{Qualitative Visualisation of the Encoded Features}
\label{sec:viz}
\begin{figure}[htbp]
    \centering
    \begin{subfigure}{\textwidth}
        \centering
        \includegraphics[width=\textwidth]{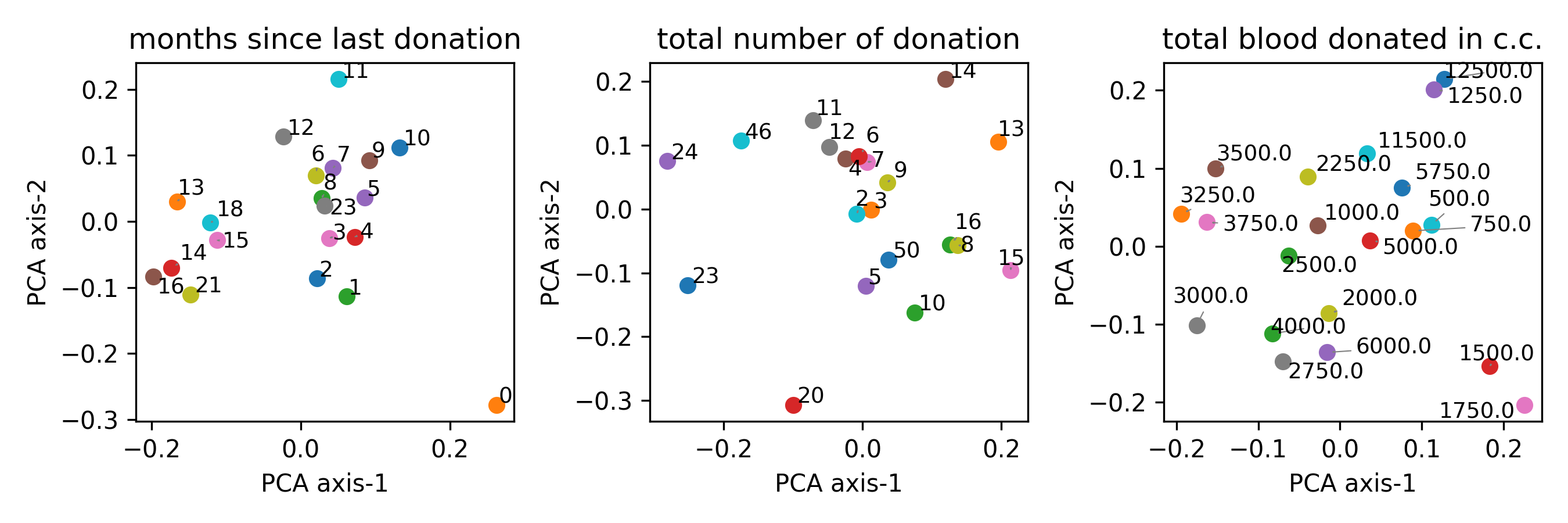}
        \caption{Visualisation of selected features for the `blood transfusion' dataset.}
        \label{fig:sub11}
    \end{subfigure}
    
    \vspace{1em} 
    
    \begin{subfigure}{\textwidth}
        \centering
        \includegraphics[width=\textwidth]{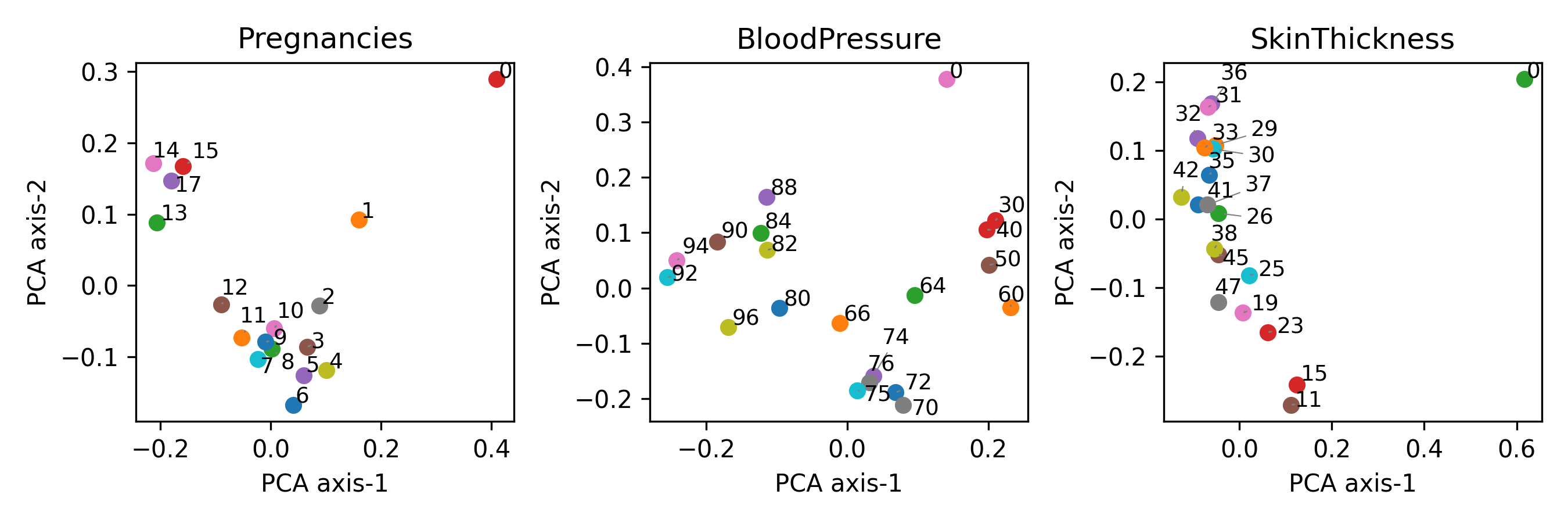}
        \caption{Visualisation of selected features for the `diabetes' dataset.}        
        \label{fig:sub22}
    \end{subfigure}

    \vspace{1em} 
    
    \begin{subfigure}{\textwidth}
        \centering
        \includegraphics[width=\textwidth]{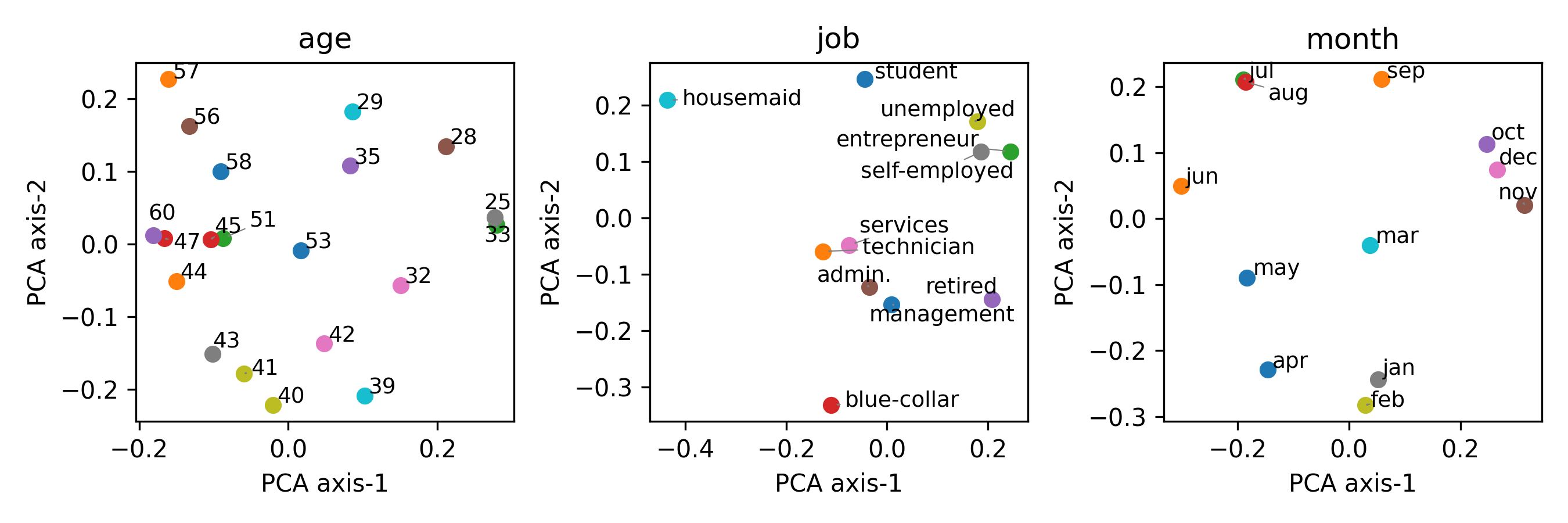}
        \caption{Visualisation of selected features for the `bank-marketing' dataset.}
        \label{fig:sub33}
    \end{subfigure}
        \begin{subfigure}{\textwidth}
        \centering
        \includegraphics[width=\textwidth]{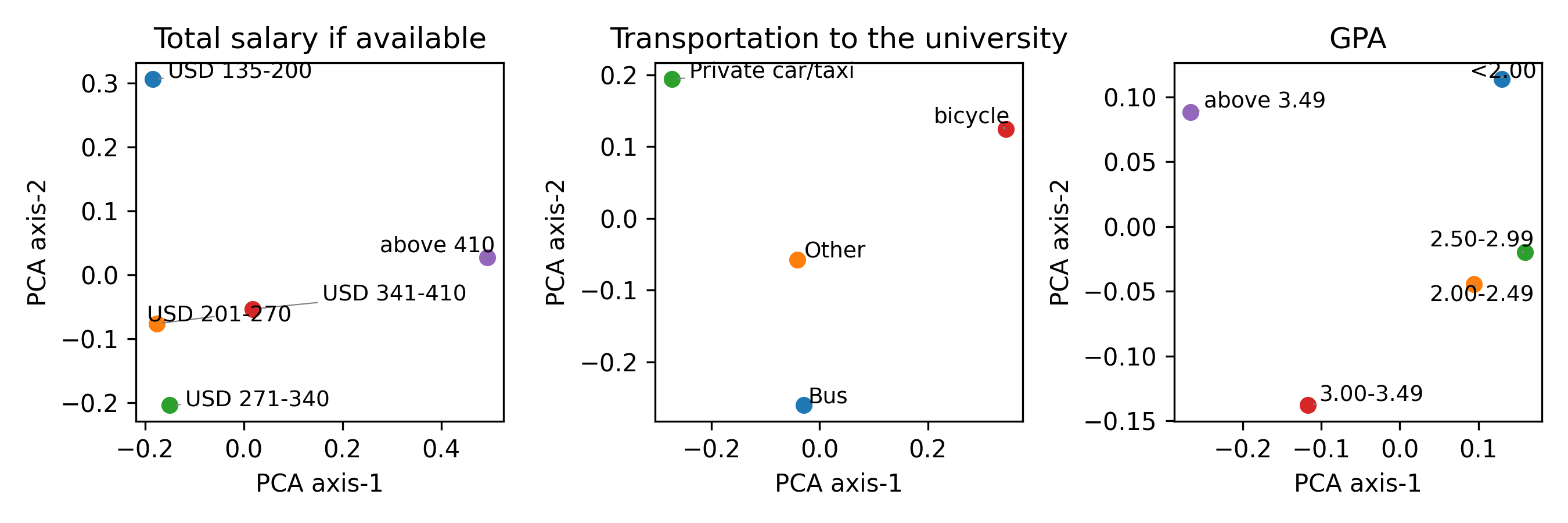}
        \caption{Visualisation of selected features for the `student-performance' dataset.}
        \label{fig:sub44}
    \end{subfigure}
    
    \caption{Projection of the embedded features with the `BGE' model. For demonstration purposes we show at most 20 randomly selected unique values.}
    \label{fig:main}
\end{figure}

In this section, we present visualisations of selected features across different datasets, check Figure \ref{fig:main}. We first embed the features using the proposed template and then project them into two dimensions via PCA. The visualisations show that both categorical and numerical features form meaningful clusters. For example, months, grade point average descriptions, and job titles group together. Interestingly, numerical values such as age, number of pregnancies, and medical measurements (e.g., total blood donated in c.c.) also cluster together. We attribute this semantic grouping to the primary influence of LLM-based embeddings on improved downstream performance. 

\clearpage

\section{Examples of Serialised Data}
\label{sec:examples}
We next show an example of serialisation for the 'students-performance' dataset (Table \ref{tab:example}) and the `blood transfusion' dataset (Table \ref{tab:example2}). These examples demonstrate that text-readable inputs enable models to better understand tabular data through natural language, which has been shown to be a significant capability of LLMs, as discussed in \cite{han2024large}.

\begin{table}[H]
    \centering
    \caption{Example of serialised row for the `students-performance' dataset.}

\resizebox{\textwidth}{!}{\begin{tabular}{lll} 
\toprule
Feature & Value & Serialised Feature:Value \\
\midrule
Student Age & 22-25 & The Student Age is 22-25. \\
Sex & male & The Sex is male. \\
Graduated high-school type & other & The Graduated high-school type is other. \\
Scholarship type & 50\% & The Scholarship type is 50\%. \\
Additional work & Yes & The Additional work is Yes. \\
Regular artistic or sports activity & No & The Regular artistic or sports activity is No. \\
Do you have a partner & No & The Do you have a partner is No. \\
Total salary if available & USD 135-200 & The Total salary if available is USD 135-200. \\
Transportation to the university & Bus & The Transportation to the university is Bus. \\
Accommodation type in Cyprus & rental & The Accommodation type in Cyprus is rental. \\
Mother's education & primary school & The Mother's education is primary school. \\
Father's education & secondary school & The Father's education is secondary school. \\
Number of sisters/brothers (if available) & 3 & The Number of sisters/brothers (if available) is 3. \\
Parental status & married & The Parental status is married. \\
Mother's occupation & housewife & The Mother's occupation is housewife. \\
Father's occupation & self-employment & The Father's occupation is self-employment. \\
Weekly study hours & 6-10 hours & The Weekly study hours is 6-10 hours. \\
Reading frequency (non-scientific books/journals) & Sometimes & The Reading frequency (non-scientific books/journals) is Sometimes. \\
Reading frequency (scientific books/journals) & Sometimes & The Reading frequency (scientific books/journals) is Sometimes. \\
Attendance to the seminars/conferences related to the department & Yes & The Attendance to the seminars/conferences related to the department is Yes. \\
Impact of your projects/activities on your success & positive & The Impact of your projects/activities on your success is positive. \\
Attendance to classes & always & The Attendance to classes is always. \\
Preparation to midterm exams 1 & alone & The Preparation to midterm exams 1 is alone. \\
Preparation to midterm exams 2 & closest date to the exam & The Preparation to midterm exams 2 is closest date to the exam. \\
Taking notes in classes & always & The Taking notes in classes is always. \\
Listening in classes & sometimes & The Listening in classes is sometimes. \\
Discussion improves my interest and success in the course & never & The Discussion improves my interest and success in the course is never. \\
Flip-classroom & useful & The Flip-classroom is useful. \\
Cumulative grade point average in the last semester (/4.00) & \textless{}2.00 & The Cumulative grade point average in the last semester (/4.00) is \textless{}2.00. \\
Expected Cumulative grade point average in the graduation (/4.00) & \textless{}2.00 & The Expected Cumulative grade point average in the graduation (/4.00) is \textless{}2.00. \\
Course ID & 1 & The Course ID is 1. \\
\bottomrule
\end{tabular}}

    \label{tab:example}
\end{table}

\begin{table}[H]

    \centering
    \caption{Example of a serialised row for the `blood transfusion' dataset.}
\begin{tabular}{lll}

\toprule
Feature & Value & Serialised Feature:Value \\
\midrule
age & 63.0 & The age is 63.0. \\
sex & 1.0 & The sex is 1.0. \\
cp & 1.0 & The cp is 1.0. \\
trestbps & 145.0 & The trestbps is 145.0. \\
chol & 233.0 & The chol is 233.0. \\
fbs & 1.0 & The fbs is 1.0. \\
restecg & 2.0 & The restecg is 2.0. \\
thalach & 150.0 & The thalach is 150.0. \\
exang & 0.0 & The exang is 0.0. \\
oldpeak & 2.3 & The oldpeak is 2.3. \\
slope & 3.0 & The slope is 3.0. \\
ca & 0.0 & The ca is 0.0. \\
thal & 6.0 & The thal is 6.0. \\
\bottomrule
\end{tabular}
    \label{tab:example2}
\end{table}

\section{Statistical Comparison}
\label{sec:stat}
\begin{figure}[H]
    \centering
    \begin{subfigure}{\textwidth}
        \centering
        \includegraphics[width=0.6\textwidth]{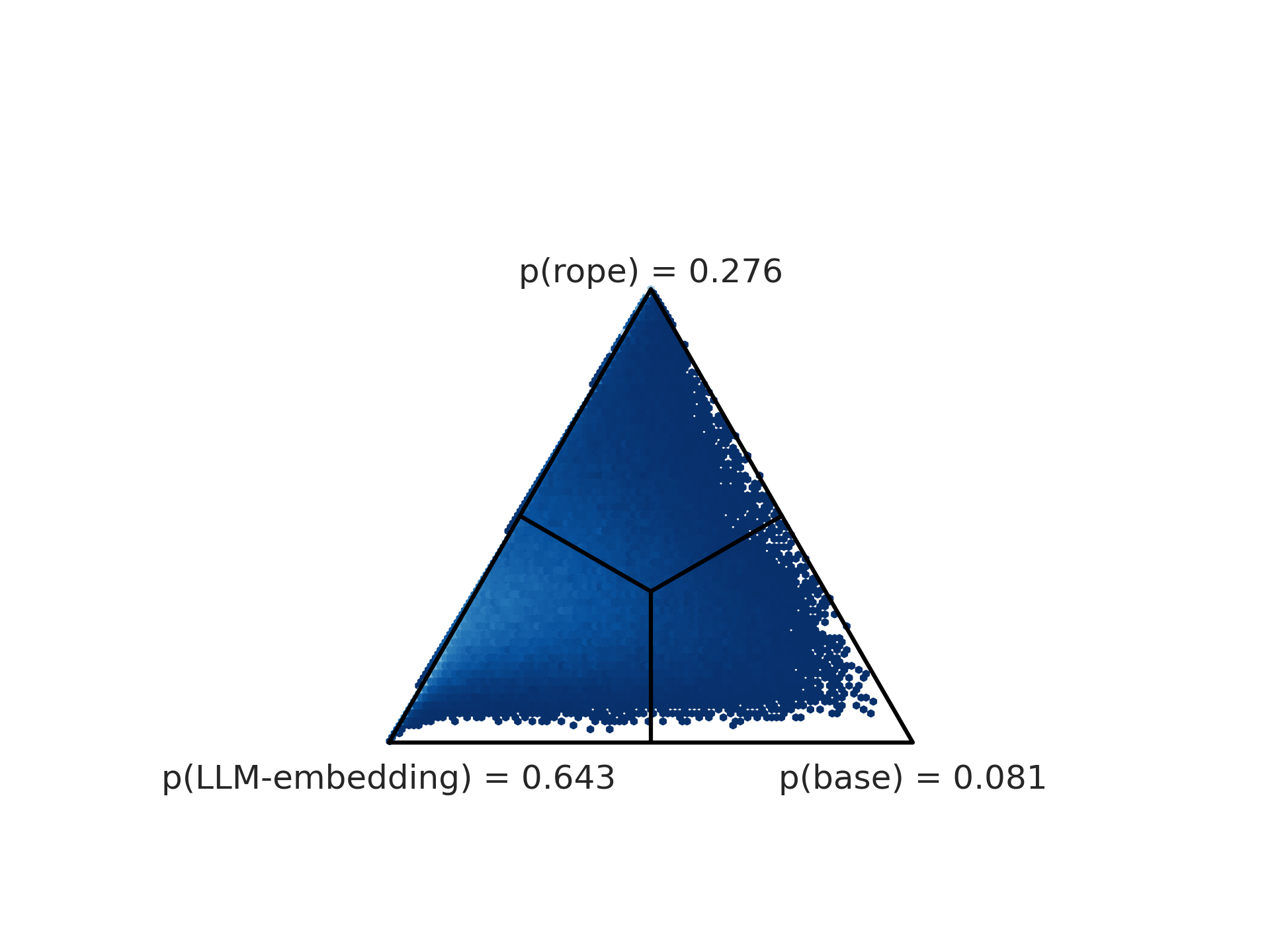}
        \caption{Visualisation of Bayesian Test for the  ResNet architecture.}
        \label{fig:sub1}
    \end{subfigure}
        
    \begin{subfigure}{\textwidth}
        \centering
        \includegraphics[width=0.6\textwidth]{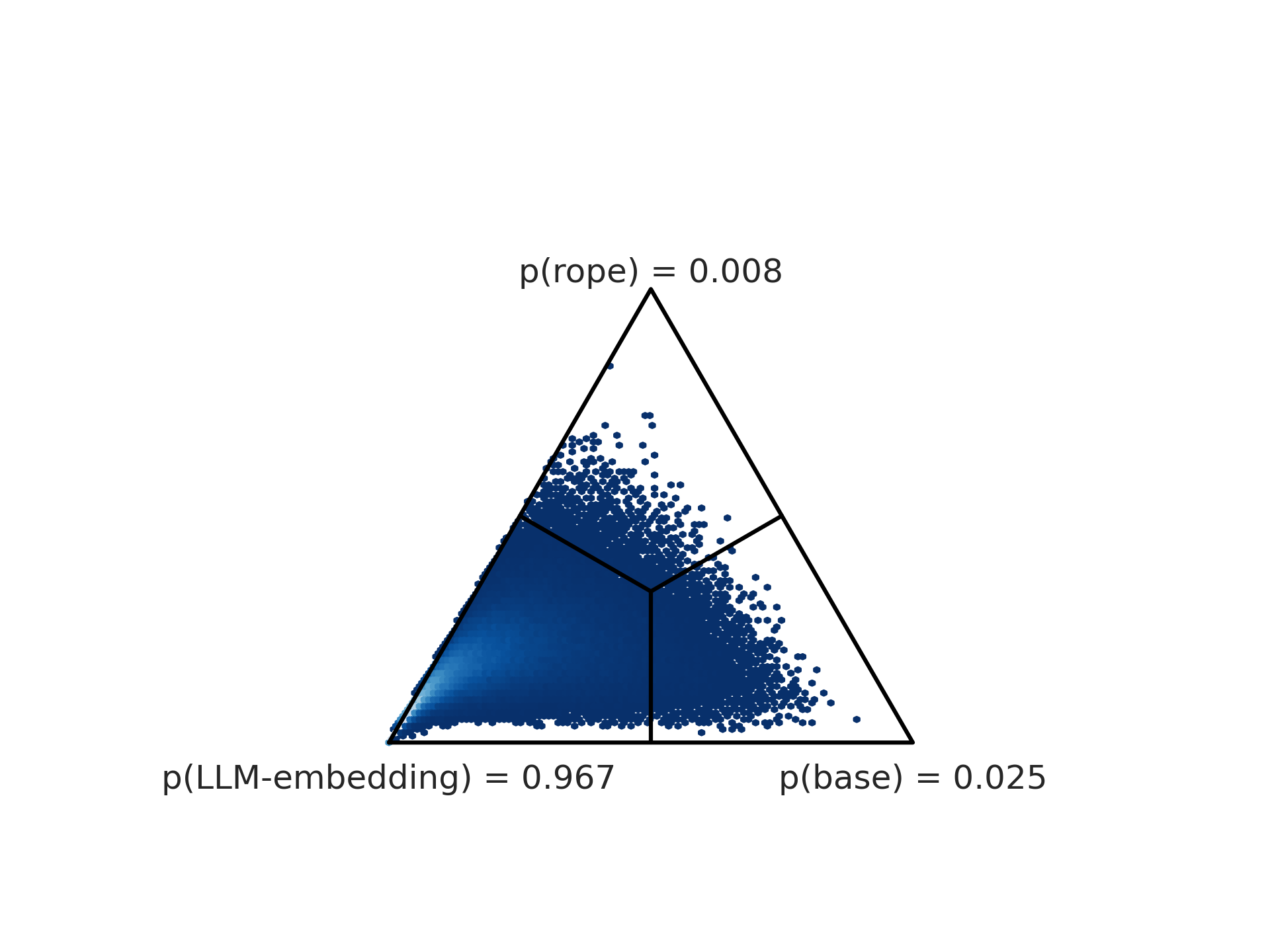}
        \caption{Visualisation of Bayesian Test for  the MLP architecture.}
        \label{fig:sub2}
    \end{subfigure}
        
    \begin{subfigure}{\textwidth}
        \centering
        \includegraphics[width=0.6\textwidth]{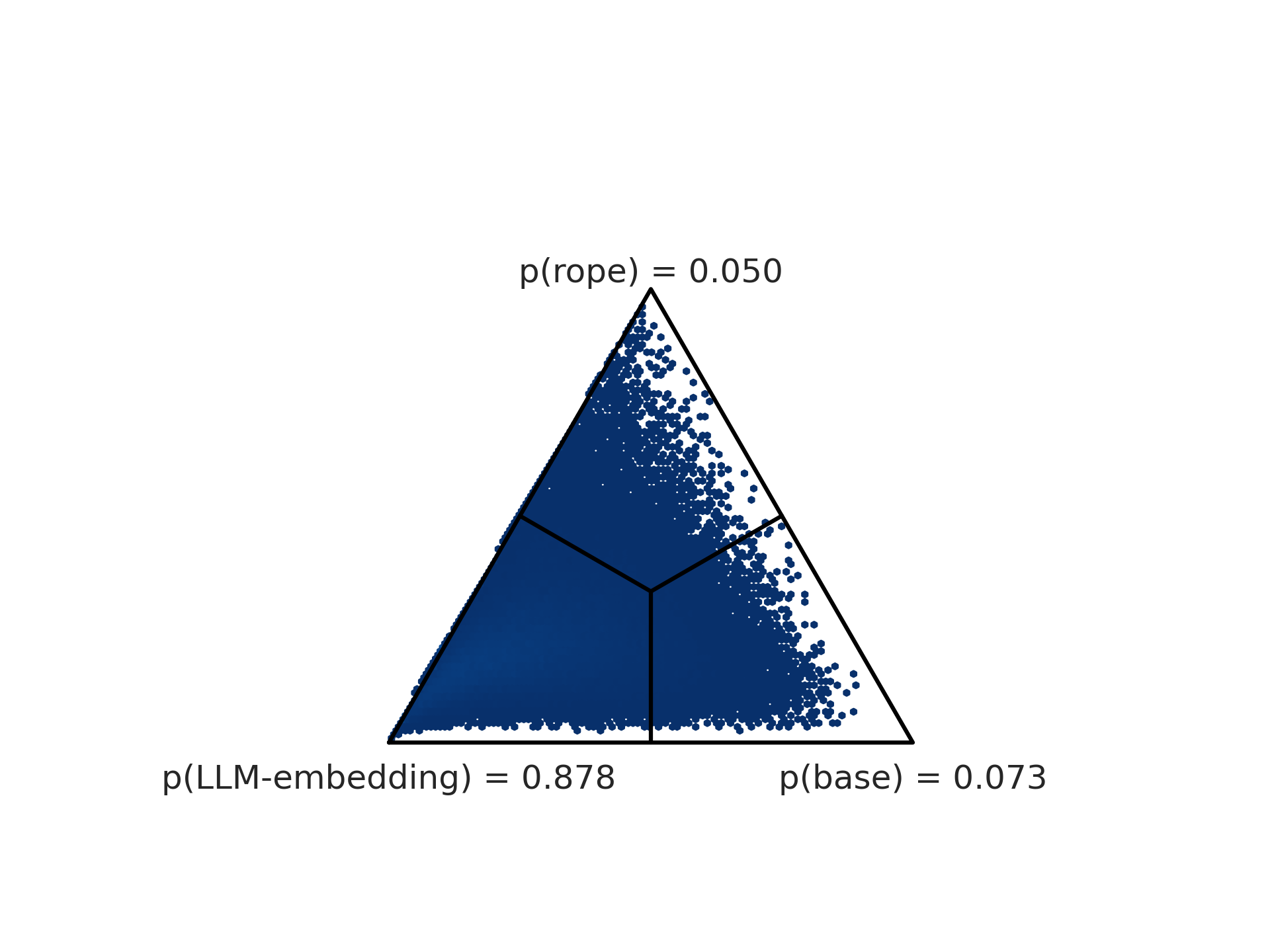}
        \caption{Visualisation of Bayesian Test for the FT-Transformer architecture.}
        \label{fig:sub3}
    \end{subfigure}
    
    \caption{Hierarchical Bayesian t-test assessing the probability that the LLM-based embeddings outperform the base embeddings across models, 7 datasets, and 10 random seeds.}
    \label{fig:bay}
\end{figure}

We conduct a hierarchical Bayesian t-test \cite{benavoli2017time} across datasets and models to assess the statistical significance of the impact of LLM-based embeddings. In each setting, we use 7 datasets and perform 10 measurements per model. Following the recommendations in \cite{benavoli2017time}, we set the region of practical equivalence to 0.1\%. Our results\footnote{When reproducing, note that results may vary non-significantly due to the Monte Carlo sampling employed in the method.} (consult Figure \ref{fig:bay}) demonstrate that our method is statistically more likely to outperform learned embeddings, supporting the applicability of our approach to current deep learning methods.

\end{document}